\documentclass[runningheads]{llncs}
\usepackage{graphicx}
\begin{document}

%
\title{A Scalable AI Approach for Clinical Trial Cohort Optimization}
%


%
\author{Xiong Liu\inst{1} \and
Cheng Shi\inst{2} \and
Uday Deore\inst{3}\and
Yingbo Wang\inst{4}\and
Myah Tran  \inst{4}\and\\
Iya Khalil\inst{1}\and
Murthy Devarakonda\inst{1}}
\authorrunning{X. Liu et al.}
%
\institute{AI Innovation Center, Novartis, Cambridge, MA, USA \and
RWE Data Science, Novartis Pharma, East Hanover, NJ, USA
\and
Global Drug Development, Novartis, East Hanover, NJ, USA\and
Global Drug Development, Novartis, Basel, Switzerland
\\
}

\maketitle              
\begin{abstract}

FDA has been promoting enrollment practices that could enhance the diversity of clinical trial populations, through broadening eligibility criteria. However, how to broaden eligibility remains a significant challenge. We propose an AI approach to Cohort Optimization (AICO) through transformer-based natural language processing of the eligibility criteria and evaluation of the criteria using real-world data. The method can extract common eligibility criteria variables from a large set of relevant trials and measure the generalizability of trial designs to real-world patients. It overcomes the scalability limits of existing manual methods and enables rapid simulation of eligibility criteria design for a disease of interest. A case study on breast cancer trial design demonstrates the utility of the method in improving trial generalizability.

\end{abstract}
\section{Introduction}

Previous studies have shown that randomized clinical trials (RCTs) are often selective and not fully representative of the real-world patients \cite{kennedy2015literature}\cite{weng2015optimizing}. Overly restrictive patient selection could lead to low enrollment and compromise study generalizability \cite{he2020clinical}. Therefore, FDA has been promoting enrollment practices that could lead to enhanced diversity in clinical trials through broadening eligibility criteria \cite{fda}.

Eligibility criteria play an essential role in defining the study population. The quality of criteria directly affects patient enrollment and study generalizability. The trial-and-error approach to define criteria often leads to many protocol amendments \cite{weng2015optimizing}. Data-driven tools are needed to help clinical trial teams discover potential patient pool that was left out before and make better eligibility criteria choices \cite{sharma2015patient}.

Clinical trial data are increasingly available through public registries (e.g., ClinicalTrials.gov). Meanwhile, real world data (RWD), such as electronic health records (EHRs), claims and billing data, are increasingly being used in drug development \cite{chen2020applications}. Combining trial designs (e.g., different eligibility criteria) with RWD could provide insights on the impact of study design in the real-world patient population and hence its generalizability post approval of the drug.

Recently, several eligibility criteria design tools based on RWD have been developed for cohort optimization. For example, Liu et al. developed a computational framework called Trial Pathfinder to systematically evaluate the effect of different eligibility criteria on cancer trial populations and outcomes using RWD \cite{liu2021evaluating}. They showed that relaxing criteria in non-small cell lung cancer trials doubled the number of eligible patients while maintaining similar hazard ratio of the overall survival in comparison to more stringent trials. Kim et al. investigated the impact of eligibility criteria on recruitment and clinical outcomes of COVID-19 clinical trials using EHR data \cite{kim2021towards}. They found that adjusting the thresholds of common eligibility criteria in COVID-19 trials could generate more outcome events with fewer patients. Similarly, Chen et al. used real-world data to simulate clinical trials of Alzheimer’s disease. They followed the study protocol of one Alzheimer's disease trial to identify the study population and considered different scenarios for simulating both intervention and control arms. The result showed that trial simulation using RWD is feasible for safety evaluation \cite{chen2021exploring}.

However, current methods still face the challenge of scalability. Trial Pathfinder by Liu et al. is based on manual encoding of the eligibility criteria, where they developed rules to encode 10 non-small cell lung cancer trials to select patients. In the study by Kim et al., two researchers manually annotated eligibility criteria from a set of 32 trials using the OMOP data model. These methods are only effective to a small number of trials and are not scalable to hundreds or even thousands of trials.

In addition, current methods focused on outcome events without an explicit measure of trial generalizability. There is a need for quantifying the study cohort representativeness with respect to the real-world patient population. Previous definitions of generalizability, such as the Generalizability Index for Study Trait (GIST) \cite{he2015simulation}, are based on the collective population representativeness of a set of trials measured by quantitative eligibility features. However, these definitions do not capture the cohort representativeness for a single study.

We propose a scalable AI framework for Cohort Optimization, thus called AICO, to automatically extract common eligibility criteria from a large set of clinical trials and evaluate the effect of these eligibility criteria on trial generalizability. Our design is based up on the recent AI advances in transformer-based natural language processing (NLP). Pre-trained language models, such as Bidirectional Encoder Representations from Transformers (BERT), have demonstrated superior performance over previous baseline models across NLP tasks such as question answering and language inference \cite{devlin2018bert}. In the clinical trial domain, a BERT-based model called CT-BERT \cite{liu2021bert} was recently introduced for named entity (or variable) extraction from the eligibility criteria. CT-BERT was built on ClinicalTrials.gov data by fine-tuning pre-trained BERT models. Previous study on eligibility criteria extraction \cite{liu2021bert} compared CT-BERT with other baseline models including the Attention-based Bidirectional LSTM (Att-BiLSTM) model \cite{liu2020attention} and the Conditional Random Field (CRF) model. The F1 scores for CT-BERT, Att-BiLSTM and CRF were 0.844, 0.802 and 0.804, respectively \cite{liu2021bert}. This shows the advantage of BERT-based NLP in eligibility criteria extraction. Getting high accuracy makes automatic AI-driven trial design feasible.

We apply CT-BERT to extract quantitative eligibility criteria variables for a set of relevant trials (up to hundreds or thousands) for a disease of interest. We then use the RWD database Optum to evaluate the generalizability of the study cohort defined by the extracted eligibility variables. We introduce a generalizability score for a single trial to measure the representativeness of the study cohort by calculating the percentage of eligible real-world patients. We also include a design capability called “what-if” analysis that allows the trial designer to adjust the thresholds of eligibility variables (e.g., clinical variables) to evaluate how different eligibility criteria designs could impact the study cohort generalizability.

We demonstrated the utility of AICO in designing the eligibility criteria of HR+ HER2- breast cancer trials. We selected an example trial with the ClinicalTrials.gov ID NCT02513394, which is a phase III double blinded, parallel-group trial. We followed the criteria of the study to identify the study population, target population and calculate the generalizability score of the trial. We adjusted the thresholds of two clinical variables of the trial: Absolute neutrophil count and Hemoglobin. The result showed that the generalizability improved by 7\% without changing other criteria.

Our contributions include: 1) automatic extraction of common eligibility criteria variables from unstructured eligibility criteria using advanced NLP; 2) identification of eligible patients from RWD (i.e., EHRs) based on the extracted eligibility criteria variables; 3) calculation of clinical trial generalizability score based on the matching between eligibility criteria and EHRs; and 4) a use case in breast cancer which shows that broadening the thresholds of clinical variables improves trial generalizability.

\section{Methods}

We propose a new approach, called AICO, to enable cohort optimization. Figure 1 shows the AICO framework, which includes BERT NLP-based extraction of eligibility variables and RWD analysis to identify cohorts and optimize trial generalizability. The details are described below.

\begin{figure}
\includegraphics[width=\textwidth]{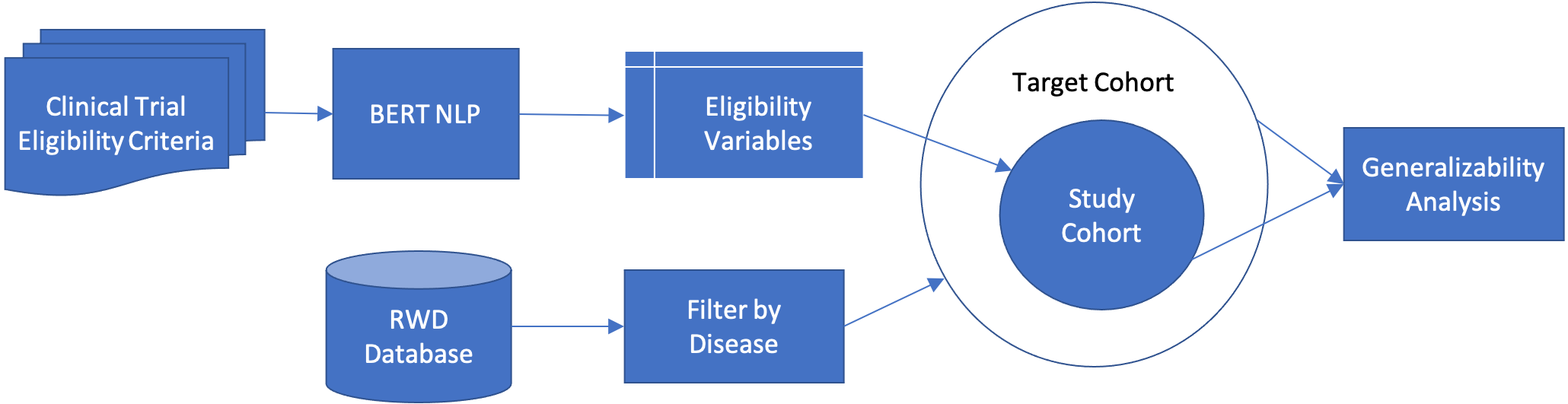}
\caption{AICO framework for generalizability analysis and cohort optimization.} \label{fig1}
\end{figure}

\subsection{Data sources}

\textbf{Clinical trial data source}: Clinical trial protocols for a disease of interest are retrieved from the ClinicalTrials.gov. The data includes metadata (e.g., title, indication, phase, year) and the eligibility criteria text.

\textbf{RWD source}: Cohort of the disease is identified and assessed from the Optum de-identified EHR database.

\subsection{NLP extraction of eligibility criteria}

Eligibility criteria are documented as unstructured text, which is not readily suitable for automated cohort definition and knowledge sharing. We apply a deep learning NLP model called CT-BERT \cite{liu2021bert} to extract variables from the eligibility criteria. CT-BERT is trained by fine-tuning pre-trained BERT models on eligibility criteria. It defines 15 entity (or variable) types in clinical trial text, such as cancer, chronic disease, treatment, and clinical variable. Previous study on a 10-trial benchmark data set shows that CT-BERT achieved F1 of 0.844 and outperformed other baseline models including attention-based BiLSTM and CRF \cite{liu2021bert}.

Given a set of trials for an indication or disease, we use CT-BERT to transform the eligibility criteria text into a set of structured eligibility variables. We then identify the most frequent variables in each variable type. These frequent variables are used to identify real-world patients from the RWD.

\subsection{Cohort identification in RWD}

From the Optum database, we identified 2 populations: the target cohort and the study cohort. The target cohort are those who will benefit from the treatment, and thus should be broader as patients with the given disease in general. The study cohort are patients who meet the eligibility criteria based on the eligibility variables in the clinical trial. 

The target cohort is identified based on the ICD10 code for the given indication. For example, in the case of metastatic breast cancer, the target cohort is identified using the ICD-10 code C50. 

The study cohort is identified by filtering the target cohort. Based on the common variables returned by the CT-BERT NLP model, we convert them to the corresponding variables available in the Optum database. The available variables in Optum are called “computable” eligibility variables, while the unavailable ones are filtered. We then use the computable eligibility variables to identify the study cohort.

\subsection{Definition of trial generalizability}

Previous definitions of trial generalizability, e.g., GIST \cite{he2015simulation}, are based on a set of trials and not applicable to individual eligibility criteria design. We define a new generalizability score for a single trial design by measuring the percentage of study cohort (\textit{SC}) among the target cohort (\textit{TC}):

\[Generalizability (T_i )=SC/TC=P(V_1,V_2,…,V_n  | I_j)/P(I_j)\] 

Where \textit{T\textsubscript{i}} is the \textit{i}th trial design, \textit{P()} is the patient count, \textit{V\textsubscript{1}}, \textit{V\textsubscript{2}}, and \textit{V\textsubscript{n}} are the computable eligibility criteria variables, and \textit{I\textsubscript{j}} is the \textit{j}th indication. If \textit{V\textsubscript{n}} is a continuous variable (e.g., age, BMI), the value range of \textit{V\textsubscript{n}} is determined by the lower bound and upper bound of the variable; if \textit{V\textsubscript{n}} is discrete (e.g., cancer stage), its value is determined by a single value.

The implementation of \textit{P()} for \textit{TC} is based on the ICD-10 code or proxy for the given indication. The detailed implementation of \textit{P()} for \textit{SC} is based on a relational patient model, where each patient is represented as a vector of the computable variables. Given a specific design, we convert the values for all the computable variables into a structured query over the patient model and retrieve the patient count.

\subsection{Evaluating the impact of criteria design}

AICO provides the capability to evaluate the criteria design for a given indication based on the computable variables. We call this capability “what-if” analysis because the user can adjust the values of the computable variables and see the generalizability score for the design. By manually adjusting the thresholds of computable variables, the user can either evaluate the generalizability of a completed trial or estimate the generalizability for a new trial design.

\section{Results}

\subsection{Breast cancer trials}

In this study we focused on the eligibility criteria design for HR+ HER2- breast cancer trials. The relevant trials were identified by search the condition or disease field of ClinicalTrials.gov using the search term ‘HR+ HER2- breast cancer’. A total of 125 trials were identified as of March 1, 2021.

\subsection{NLP extraction of eligibility criteria}

Using CT-BERT, we extracted 11,709 entities from 3,572 criteria sentences in the 125 breast cancer trials. Table 1 shows some examples of the extracted entities and associated entity types. CT-BERT can recognize complex entity types with high accuracy. For example, it can extract “hemoglobin” as clinical variable and “9g/dl” as its lower bound in a simple criterion (1st criterion). It can also extract multiple variables such as “absolute neutrophil count”, “platelets”, and “total bilirubin” and associated value ranges in more complex criterion (2nd criterion). Other variables such as prior disease and treatment are also extracted. The extracted variable names are standardized using a rule-based method to facilitate the follow-up analysis. For example, “ANC” and “absolute neutrophil count (ANC)” are standardized to the same concept of “absolute neutrophil count”.

\begin{table}[t]
\caption{Examples of extracted entities from the eligibility criteria of breast cancer trials}
\label{table2:NERexamples}
\begin{tabular}{lp{5cm}|p{3cm}|l}

\multicolumn{1}{c}{\bf Eligibility}   &\multicolumn{1}{c}{\bf Criterion}  &\multicolumn{1}{c}{\bf Entity} &\multicolumn{1}{c}{\bf Type}
\\ \hline \\
inclusion  &Hemoglobin \textgreater= 9 g/dL (90 g/L) &Hemoglobin &clinical variable \\
\cline{3-4}
& &9 g/dL (90 g/L) &lower bound \\
\\ \hline \\
inclusion  &absolute neutrophil count \textgreater= 1,500/mcL; platelets \textgreater= 100,000/mcL; total bilirubin within 1.25 x normal institutional limits  &absolute neutrophil count &clinical variable \\
\cline{3-4}
& &1,500/mcL &lower bound \\
\cline{3-4}
& &platelets &clinical variable \\
\cline{3-4}
& &100,000/mcL &lower bound \\
\cline{3-4}
& &total bilirubin &clinical variable \\
\cline{3-4}
& &1.25 x normal institutional limits &upper bound\\
\\ \hline \\
exclusion &History of liver disease, such as cirrhosis or chronic active hepatitis B and C &liver disease &disease\\
\cline{3-4}
& &cirrhosis &disease \\
\cline{3-4}
& &chronic active hepatitis b and c &disease\\
\\ \hline \\
exclusion &Food or drugs that are known to be CYP3A4 inhibitors &CYP3A4 inhibitors &treatment\\
\end{tabular}
\end{table}

\subsection{Common eligibility criteria in breast cancer trials}

We identified common eligibility criteria variables based on their frequency in the 125 trials. The most frequent variables across all types include demographic variables (e.g., women, age), disease variables (e.g., breast cancer, metastatic disease), treatment variables (e.g., chemotherapy, endocrine therapy), and clinical variables (e.g., ECOG, platelets, creatinine).

Figure 2 shows the most frequent clinical variables ranked by the number of unique trials. The top 5 variables include ECOG, platelets, creatinine, bilirubin, and hemoglobin. 85 trials considered ECOG in the eligibility criteria, making it the most common clinical variable.

\begin{figure}
\includegraphics[width=\textwidth]{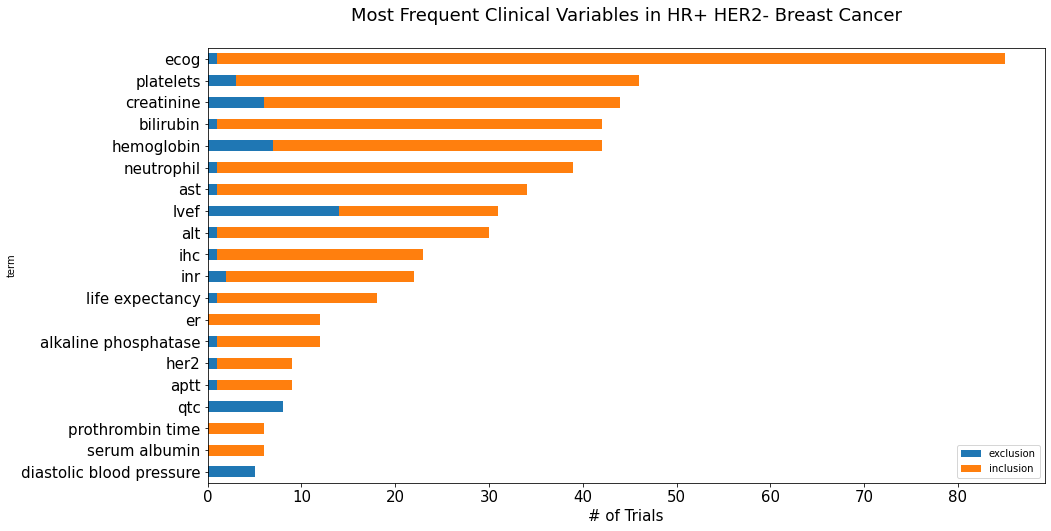}
\caption{Top 20 clinical variables in the eligibility criteria of breast cancer trials.} \label{fig1}
\end{figure}

\subsection{The target cohort and study cohort in RWD}

We identified the target cohort of HR+ HER2- breast cancer patients based on ICD-10 code C50. We did not consider sub-cohort by HR or HER2 status because it is not available in ICD-10. A better method to estimate HR and HER2 status \cite{twelves2020systemic} shall be developed in the future. In addition, we used the overall index period as 2009 – 2021 based on the start dates of the 125 trials. In total, we identified 199,169 patients in the Optum database during the overall index period.

For simplification, we used the most common clinical variables to define the study cohort. We then identified the ‘computable’ variables available in Optum, including ALT, AST, bilirubin, creatinine, hemoglobin, and neutrophil. All the values for the variables were cleaned and normalized with the standard unit. The same set of variables were also used to characterize the target cohort.

\subsection{Cohort optimization study}

We studied the impact of the thresholds of the clinical variables on generalizability for three clinical trials, including one real trial (NCT02513394) and two simulated trials. 

The NCT02513394 trial is a Phase III study for patients with HR+ / HER2- early breast cancer (EBC). It has the following inclusion criteria using 4 common clinical variables: 1) Absolute neutrophil count \textgreater= 1.5 109/L; 2) Hemoglobin \textgreater= 10 g/dl; 3) Total Bilirubin \textless= 3 x ULN; and 4) AST \textless= 1.5 x ULN. We selected the ULN for total bilirubin as 1.2 mg/dL and the ULN for AST as 24 units per liter. The enrollment start date was Aug 26, 2015 and end date Feb 11, 2019. We selected the index period for real-world patients as Jan 2014 to Jan 2019 such that the last diagnosis date is within 2 year of the enrollment start date and before the enrollment end date of the trial.

The simulated trials are based on the patient event grade definitions in the NCI Common Terminology Criteria for Adverse Events \cite{bworld}, where Grade 1 patients have a lower bound of absolute neutrophil as 1.5 x 109/L and a lower bound of Hemoglobin as 10 g/dl, and Grade 2 patients have a lower bound of absolute neutrophil as 1.0 x 109/L and a lower bound of Hemoglobin as 8 g/dl. The 1st simulated trial modified NCT02513394 by broadening the lower bound of absolute neutrophil to 1.0 x 109/L (Grade 2). The 2nd simulated trial broadened the lower bound of absolute neutrophil to 1.0 x 109/L (Grade 2) and the lower bound of Hemoglobin to 8 g/dl (Grade 2). Both simulations used the same index period as in NCT02513394.

Table 2 shows the number of eligible patients (study cohort), the number of total HR+ HER2- patients during the index period (target cohort), and the generalizability score (percentage of study cohort among target cohort) for all trials. The real trial has a generalizability score of 80.15\%, the 1st simulated trial improved the generalizability to 80.90\% by broadening only absolute neutrophil, and the 2nd simulated trial further improved the generalizability to 85.70\% by reducing the hemoglobin level threshold. This demonstrates that relaxing the thresholds of clinical variables can improve the trial generalizability up to 7\% (80.15\% vs 85.70\%).

\begin{table}[h]
\caption{The number of patients and trial generalizability in breast cancer trials}
\label{table2:NERexamples}
\begin{center}
\begin{tabular}{p{0.2\textwidth}|p{0.41\textwidth}|p{0.1\textwidth}|p{0.1\textwidth}|p{0.1\textwidth}}
\multicolumn{1}{c}{\bf Trial}   &\multicolumn{1}{c}{\bf Eligibility Criteria}   
&\bf Study Cohort
&\bf Target Cohort
&\bf Generalizability
\\ \hline \\
Real trial (NCT02513394)     &•	Absolute neutrophil $\geq$ 1.5 109/L  &75,432 &94,114 &80.15\% \\
&•	Hemoglobin $\geq$ 10 g/dl & & &\\
&•	Total Bilirubin $\leq$ 3 x ULN & & &\\
&•	AST $\leq$ 1.5 x ULN & & &\\
\\ \hline \\
Simulated trial 1 &•	Absolute neutrophil $\geq$ 1.0 109/L &76,132 &94,114 &80.90\% \\
&•	Hemoglobin $\geq$ 10 g/dl & & &\\
&•	Total Bilirubin $\leq$ 3 x ULN & & &\\
&•	AST $\leq$ 1.5 x ULN & & &\\
\\ \hline \\
Simulated trial 2 &•	Absolute neutrophil $\geq$ 1.0 109/L &80,655 &94,114 &85.70\% \\
&•	Hemoglobin $\geq$ 8 g/dl & & &\\
&•	Total Bilirubin $\leq$ 3 x ULN & & &\\
&•	AST $\leq$ 1.5 x ULN & & &\\

\end{tabular}
\end{center}
\end{table}

\section{Discussion}

\subsection{AICO advantages}

AICO has several distinctive features. First, it can automatically process a large number of relevant trials to identify common eligibility criteria variables. This provides improvement over other methods which are manual and limited to a small number of trials. Although manual extraction of eligibility criteria ensures accuracy, it faces a scalability issue at the same time. Our method leverages the cutting-edge BERT-based models (i.e., CT-BERT) to extract eligibility criteria variables. Previous study has shown that CT-BERT achieved a F1 score of 0.844 and outperformed other baselines based on a 10-trial benchmark data set. We did not measure the F1 score for the current breast cancer study due to the lack of benchmark data. Visual inspection of the extraction results showed that their quality is sufficient to enable downstream RWD analysis. In fact, the common eligibility variables shown in Figure 2 confirmed our prior knowledge about breast cancer trials.

Second, AICO defines an intuitive generalizability score for each trial design by comparing the study cohort with the target cohort in RWD. Previous cohort optimization methods focused on evaluating outcome events instead of measuring the trial generalizability. Also, previous generalizability studies defined generalizability for a set of trials, which is not applicable to a single trial design. Our generalizability definition not only connects the common eligibility variables with the RWD patients but also measures the representativeness of study cohort.

Third, AICO enables cohort optimization through what-if analysis. Our case study in breast cancer shows that adjusting the thresholds of clinical variables improved the generalizability. Although the threshold adjustment is still a manual process, it is possible to incorporate domain knowledge about the trials to make more meaningful adjustment. For example, we adjusted the lower bounds of absolute neutrophil and hemoglobin for NCT02513394 (a phase III trial for breast cancer) based on the patient event grades defined in the NCI Common Terminology Criteria for Adverse Events \cite{bworld}. So the adjusted thresholds should not be totally arbitrary.

\subsection{Limitations}

This study also has several limitations. First, we leveraged the existing entity or variable type definition in CT-BERT without considering disease or trial specific entities. CT-BERT includes 15 entity types in clinical text such as disease, treatment, clinical variables, and value ranges. However, more fine-grained types such as mutation status, biopsy and histology are still needed for oncology trials.

Second, the study cohort definition is based on common eligibility criteria variables across relevant trials. Less common or frequent variable types were not used, which could miss unique features for individual trials. However, it is possible to combine common variables and trial-specific variables to better define the cohort for a specific trial.

Third, we only performed case studies using the Optum database. Although Optum provides comprehensive EHRs, it may not include all variables identified by the CT-BERT NLP model. That is why we introduced ‘computable’ variables to only include available variables in Optum RWD. To enable more computable variables, more RWD databases such as Flatiron will need to be explored.

\section{Conclusions}

We present AICO, a new framework for scalable cohort optimization based on advanced NLP and RWD assessment of trial generalizability. It facilitates eligibility criteria design for a given clinical trial of interest, allowing possibility for the study to enroll faster and to enroll patients who are most likely to receive the drug if approved. We performed a case study in HR+ HER2- breast cancer to automatically extract eligibility variables from hundreds of relevant trials to enable cohort definition and identification. We then experimented how different eligibility criteria choices may impact the size of the study cohort and the trial generalizability. We found that broadening the thresholds of clinical variables could improve the generalizability through trial simulations. Thus, our approach overcomes the scalability limits of previous manual methods and enables rapid simulation of trial design towards better generalizability.

Future work will include 1) identifying and extracting new variable types that are not covered by existing NLP models, 2) refining study cohort definition to include both common and trial-specific variables, 3) strategies to map eligibility variables to more ‘computable’ variables in RWD by exploring more RWD databases, and 4) developing more use cases in different disease areas.


\typeout{}
\bibliography{refs}

\bibliographystyle{splncs04}

\end{document}